%% file: main.tex
\documentclass[letterpaper, 10 pt, journal, twoside]{IEEEtran}
\usepackage{framed,multirow}
\usepackage{amssymb}
\usepackage{url}
\usepackage[dvipsnames]{xcolor}
\usepackage{colortbl}

\usepackage{amsmath}
\usepackage{latexsym}
\usepackage{booktabs}
\usepackage{graphics}
\usepackage{graphicx}
\usepackage{rotating}
\usepackage{duckuments}
\usepackage{tablefootnote}

\newcommand{\etal}{\textit{et al}.}
\newcommand{\ie}{\textit{i}.\textit{e}.}
\newcommand{\eg}{\textit{e}.\textit{g}.}
\newcommand{\tit}[1]{\smallbreak\noindent\textbf{#1.}}

\newcommand{\tinytit}[1]{\noindent\textbf{#1.}}
\newcommand{\tinytextit}[1]{\noindent\textit{#1:}}

\newcommand{\rev}[1]{\textcolor{black}{#1}}

\begin{document}
\title{Focus on Impact:\\Indoor Exploration with Intrinsic Motivation}

\author{\IEEEauthorblockN{Roberto Bigazzi, Federico Landi, Silvia Cascianelli, Lorenzo Baraldi, Marcella Cornia, Rita Cucchiara\\}
\thanks{© 2022 IEEE. Personal use of this material is permitted. Permission from IEEE must be obtained for all other uses, in any current or future media, including reprinting/republishing this material for advertising or promotional purposes, creating new collective works, for resale or redistribution to servers or lists, or reuse of any copyrighted component of this work in other works.
This work was supported by the ``European Training Network on PErsonalized Robotics as SErvice Oriented applications'' (PERSEO) MSCA-ITN-2020 project (G.A. 955778).} %Use only for final RAL version
\thanks{The authors are with the Department of Engineering ``Enzo Ferrari'', University of Modena and Reggio Emilia, Italy
(e-mail: {\tt\footnotesize \{name.surname\}@unimore.it})}%
}

\markboth{IEEE Robotics and Automation Letters. Preprint Version. Accepted January, 2022}
{Bigazzi \MakeLowercase{\textit{et al.}}: Focus on Impact}


\maketitle

\begin{abstract}
Exploration of indoor environments has recently experienced a significant interest, also thanks to the introduction of deep neural agents built in a hierarchical fashion and trained with Deep Reinforcement Learning (DRL) on simulated environments. Current state-of-the-art methods employ a dense extrinsic reward that requires the complete a priori knowledge of the layout of the training environment to learn an effective exploration policy. However, such information is expensive to gather in terms of time and resources. \rev{In this work, we propose to train the model with a purely intrinsic reward signal to guide exploration, which is based on the impact of the robot's actions on its internal representation of the environment.} So far, impact-based rewards have been employed for simple tasks and in procedurally generated synthetic environments with countable states. Since the number of states observable by the agent in realistic indoor environments is non-countable, we include a neural-based density model and replace the traditional count-based regularization with an estimated pseudo-count of previously visited states. The proposed exploration approach outperforms DRL-based competitors relying on intrinsic rewards and surpasses the agents trained with a dense extrinsic reward computed with the environment layouts. We also show that a robot equipped with the proposed approach seamlessly adapts to point-goal navigation and real-world deployment.
\end{abstract}

\input{sections/01-intro}
\input{sections/02-related}
\input{sections/03-method}
\input{sections/04-experiments}
\input{sections/05-conclusion}

\bibliographystyle{IEEEtran}
\bibliography{bibliography.bib}

\end{document}

%% file: sections/01-intro.tex
\section{Introduction}
\label{sec:introduction}
\IEEEPARstart{R}{obotic} exploration is the task of autonomously navigating an unknown environment with the goal of gathering sufficient information to represent it, often via a spatial map~\cite{stachniss2009robotic}. This ability is key to enable many downstream tasks such as planning~\cite{selin2019efficient} and goal-driven navigation~\cite{savva2019habitat,wijmans2019dd,morad2021embodied}. 
Although a vast portion of existing literature tackles this problem~\cite{osswald2016speeding,chaplot2019learning,chen2019learning,ramakrishnan2020occupancy}, it is not yet completely solved, especially in complex indoor environments.
The recent introduction of large datasets of photorealistic indoor environments~\cite{chang2017matterport3d,xia2018gibson} has eased the development of robust exploration strategies, which can be validated safely and quickly thanks to powerful simulating platforms~\cite{deitke2020robothor,savva2019habitat}. Moreover, exploration algorithms developed on simulated environments can be deployed in the real world with little hyperparameter tuning~\cite{kadian2020sim2real,bigazzi2021out,truong2021bi}, if the simulation is sufficiently realistic. 

Most of the recently devised exploration algorithms exploit deep reinforcement learning (DRL)~\cite{zhu2018deep}, as learning-based exploration and navigation algorithms are more flexible and robust to noise than geometric methods~\cite{chen2019learning,niroui2019deep,ramakrishnan2021exploration}. 
Despite these advantages, one of the main challenges in training DRL-based exploration algorithms is designing appropriate rewards. 
\begin{figure}[t!]
\centering
\includegraphics[width=0.9\linewidth]{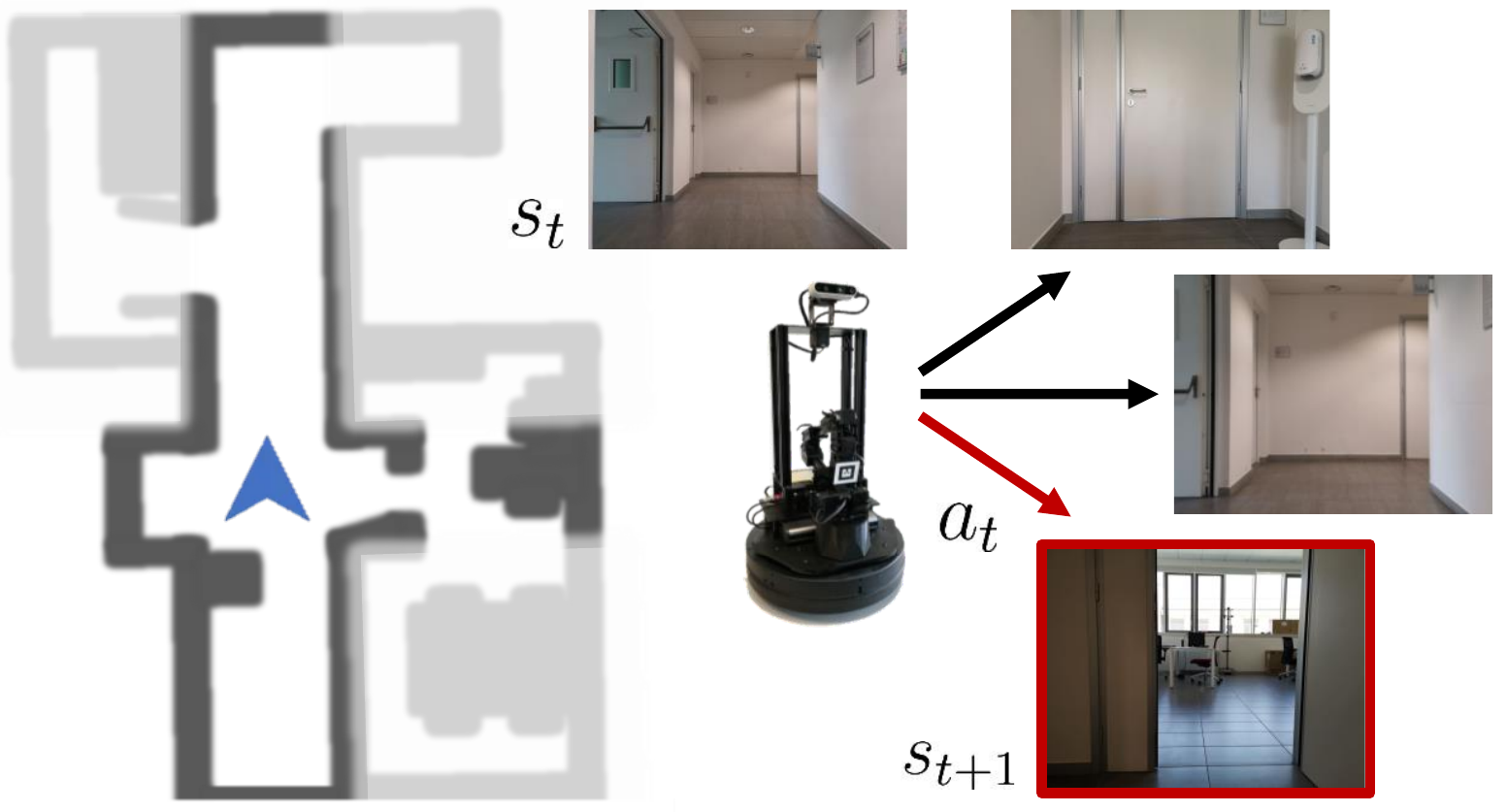}
\caption{We propose an impact-based reward for robot exploration of continuous indoor spaces. The robot is encouraged to take actions that maximize the difference between two consecutive observations.}
\label{fig:overview}
\vspace{-0.25cm}
\end{figure}
%
% Novelty anticipation
%
In this work, we propose a new reward function that employs the impact of the agent actions on the environment, measured as the difference between two consecutive observations~\cite{raileanu2020ride}, discounted with a pseudo-count~\cite{bellemare2016unifying} for previously-visited states (see Fig~\ref{fig:overview}).
So far, impact-based rewards~\cite{raileanu2020ride} have been used only as an additional intrinsic reward in procedurally-generated (\eg~grid-like mazes) or singleton (\ie~the test environment is the same employed for training) synthetic environments.
Instead, our reward can deal with photorealistic non-singleton environments. To the best of our knowledge, this is the first work to apply impact-based rewards to this setting. 

Recent research on robot exploration proposes the use of an extrinsic reward based on occupancy anticipation~\cite{ramakrishnan2020occupancy}.
This reward encourages the agent to navigate towards areas that can be easily mapped without errors. \rev{Unfortunately, this approach presents a major drawback, as this reward heavily depends on the mapping phase, rather than focusing on what has been already seen. In fact, moving towards new places that are difficult to map would produce a very low occupancy-based reward. Moreover, the precise layout of the training environments is not always available, especially in real-world applications.}
To overcome these issues, a different line of work focuses on the design of intrinsic reward functions, that can be computed by the agent by means of their current and past observations. Some examples of recently proposed intrinsic rewards for robot exploration are based on curiosity~\cite{bigazzi2020explore}, novelty~\cite{ramakrishnan2021exploration}, and coverage~\cite{chaplot2019learning}. All these rewards, however, tend to vanish with the length of the episode because the agent quickly learns to model the environment dynamics and appearance (for curiosity and novelty-based rewards) or tends to stay in previously-explored areas (for the coverage reward). Impact, instead, provides a stable reward signal throughout all the episode~\cite{raileanu2020ride}. 

Since robot exploration takes place in complex and realistic environments that can present an infinite number of states, it is impossible to store a visitation count for every state. Furthermore, the vector of visitation counts would consist of a very sparse vector, and that would cause the agent to give the same impact score to nearly identical states. To overcome this issue, we introduce an additional module in our design to keep track of a pseudo-count for visited states. The pseudo-count is estimated by a density model trained end-to-end and together with the policy.
We integrate our newly-proposed reward in a modular embodied exploration and navigation system inspired by that proposed by Chaplot~\etal~\cite{chaplot2019learning} and consider two commonly adopted collections of photorealistic simulated indoor environments, namely Gibson~\cite{xia2018gibson} and Matterport 3D (MP3D)~\cite{chang2017matterport3d}. Furthermore, we also deploy the devised algorithm in the real world. The results in both simulated and real environments are promising: we outperform state-of-the-art baselines in simulated experiments and demonstrate the effectiveness of our approach in real-world experiments. \rev{We make the source code of our approach and pre-trained models publicly available\footnote{\rev{\url{https://github.com/aimagelab/focus-on-impact}}}.}

%% file: sections/02-related.tex
\section{Related Work}
\label{sec:related}
\tinytit{Geometric Robot Exploration Methods}
Classical heuristic and geometric-based exploration methods rely on two main strategies: frontier-based exploration~\cite{yamauchi1997frontier} and next-best-view planning~\cite{gonzalez2002navigation}.
These methods have been largely used and improved~\cite{holz2010evaluating,bircher2016receding,niroui2019deep} or combined in a hierarchical exploration algorithm~\cite{zhu2018deep,selin2019efficient}. However, when applied with noisy odometry and localization sensors or in highly complex environments, geometric approaches tend to fail~\cite{chen2019learning,niroui2019deep,ramakrishnan2021exploration}. In light of this, increasing research effort has been dedicated to the development of learning-based approaches, which usually exploit DLR to learn robust and efficient exploration policies.

\tit{Intrinsic Exploration Rewards} The lack of ground-truth in the exploration task forces the adoption of reinforcement learning (RL) for training exploration methods. 
Even when applied to tasks different from robot exploration, RL methods have low sample efficiency. Thus, they require designing intrinsic reward functions that encourage visiting novel states or learning the environment dynamics. The use of intrinsic motivation is beneficial when external task-specific rewards are sparse or absent.
Among the intrinsic rewards that motivate the exploration of novel states, Bellemare~\etal~\cite{bellemare2016unifying} introduced the notion of pseudo visitation count by using a Context-Tree Switching (CTS) density model to extract a pseudo-count from raw pixels and applied count-based algorithms. Similarly, Ostrovski~\etal~\cite{ostrovski2017count} applied the autoregressive deep generative model PixelCNN~\cite{oord2016conditional} to estimate the pseudo-count of the visited state.
\rev{Recently, Zhang \etal~\cite{zhang2020bebold} proposed a criterion to mitigate common issues in count-based methods.}
Rewards that encourage the learning of the environment dynamics comprehend Curiosity~\cite{pathak2017curiosity}, Random Network Distillation (RND)~\cite{burda2018exploration}, and Disagreement~\cite{pathak2019self}.
Recently, Raileanu~\etal~\cite{raileanu2020ride} proposed to jointly encourage both the visitation of novel states and the learning of the environment dynamics. However, their approach is developed for grid-like environments with a finite number of states, where the visitation count can be easily employed as a discount factor. In this work, we improve Impact, a paradigm that rewards the agent proportionally to the change in the state representation caused by its actions, and design a reward function that can deal with photorealistic scenes with non-countable states.

\tit{Learning-based Robot Exploration Methods}
In the context of robot exploration and navigation tasks, the introduction of photorealistic simulators has represented a breeding ground for the development of self-supervised DRL-based visual exploration methods. Ramakrishnan~\etal~\cite{ramakrishnan2021exploration} identified four paradigms for visual exploration: novelty-based, curiosity-based (as defined above), reconstruction-based, and coverage-based.
Each paradigm is characterized by a different reward function used as a self-supervision signal for optimizing the exploration policy.
A coverage-based reward, considering the area seen, is also used in the modular approach to Active SLAM presented in~\cite{chaplot2019learning}, which combines a neural mapper module with a hierarchical navigation policy. To enhance exploration efficiency in complex environments, Ramakrishnan~\etal~\cite{ramakrishnan2020occupancy} resorted to an extrinsic reward by introducing the occupancy anticipation reward, which aims to maximize the agent accuracy in predicting occluded unseen areas.

\begin{figure*}[t!]
\centering
\includegraphics[width=0.98\textwidth]{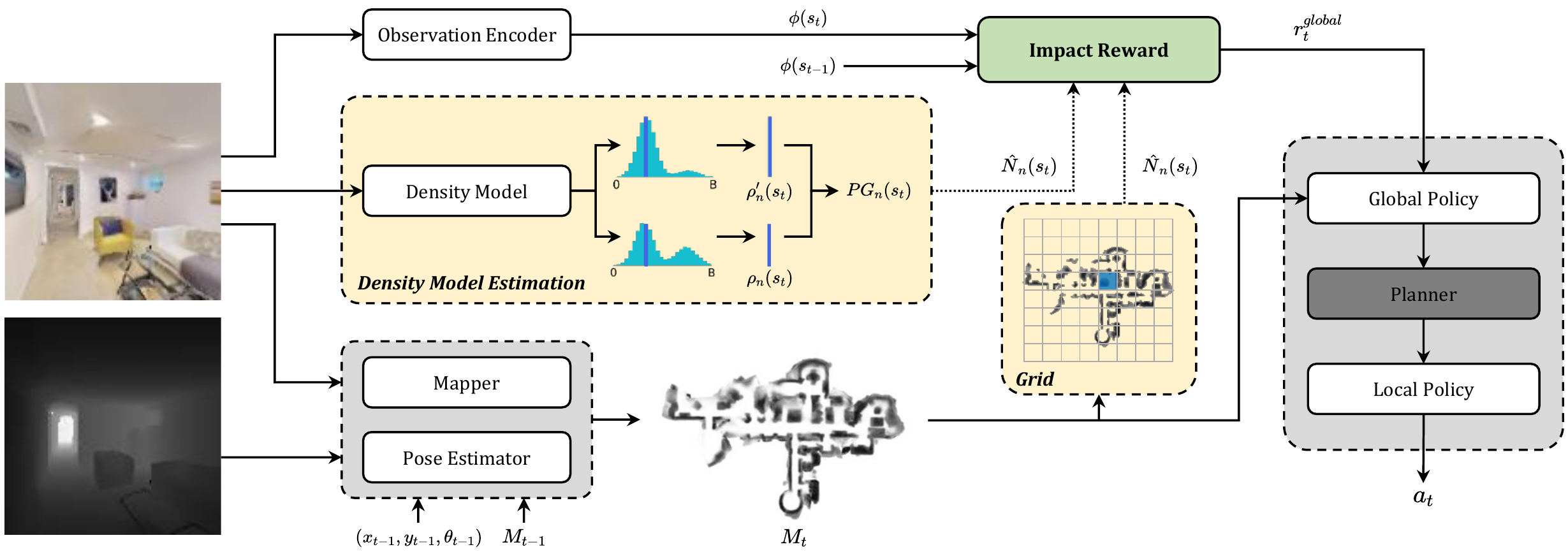}
\caption{Our modular exploration architecture consists of a Mapper that iteratively builds a top-down occupancy map of the environment, a Pose Estimator that predicts the pose of the robot at every step, and a hierarchical self-supervised Navigation Module in charge of sequentially setting exploration goals and predicting actions to navigate towards it. We exploit the impact-based reward to guide the exploration and adapt it for continuous environments, using an Observation Encoder to extract observation features and depending on the method, a Density Model or a Grid to compute the pseudo-count.}
\label{fig:method}
\vspace{-0.25cm}
\end{figure*}

\tit{Deep Generative Models} Deep generative models are trained to approximate high-dimensional probability distributions by means of a large set of training samples. In recent years, literature on deep generative models followed three main approaches: latent variable models like VAE~\cite{kingma2013auto}, implicit generative models like GANs~\cite{goodfellow2014generative}, and exact likelihood models. 
Exact likelihood models can be classified in non-autoregressive flow-based models, like RealNVP~\cite{dinh2016density} and Flow++~\cite{ho2019flow++}, and autoregressive models, like PixelCNN~\cite{oord2016conditional} and Image Transformer~\cite{parmar2018image}. Non-autoregressive flow-based models consist of a sequence of invertible transformation functions to compose a complex distribution modeling the training data. Autoregressive models decompose the joint distribution of images as a product of conditional probabilities of the single pixels. Usually, each pixel is computed using as input only the previous predicted ones, following a raster scan order. In this work, we employ PixelCNN~\cite{oord2016conditional} to learn a probability distribution over possible states and estimate a pseudo visitation count.

%% file: sections/03-method.tex
\section{Proposed Method}
\label{sec:method}

\subsection{Exploration Architecture}
\rev{Following the current state-of-the-art architectures for navigation for embodied agents~\cite{chaplot2019learning,ramakrishnan2020occupancy}, the proposed method comprises three main components: a CNN-based mapper, a pose estimator, and a hierarchical navigation policy.} The navigation policy defines the actions of the agent, the mapper builds a top-down map of the environment to be used for navigation, and the pose estimator locates the position of the agent on the map. Our architecture is depicted in Fig.~\ref{fig:method} and described below.

\subsubsection{Mapper}
The mapper generates a map of the free and occupied regions of the environment discovered during the exploration. 
\rev{At each time step, the RGB observation $o^{rgb}_t$ and the depth observation $o^{d}_t$ are processed to output a two-channel $V\times V$ local map $l_t$ depicting the area in front of the agent, where each cell describes the state of a $5\times 5$ cm area of the environment, the channels measure the probability of a cell being occupied and being explored, as in~\cite{chaplot2019learning}.}
\rev{Please note that this module performs anticipation-based mapping, as defined in~\cite{ramakrishnan2020occupancy}, where the predicted local map $l_t$ includes also unseen/occluded portions of space.}
The local maps are aggregated and registered to the $W\times W\times 2$ global map $M_t$ of the environment using the estimated pose $(x_t, y_t, \theta_t)$ from the pose estimator. The resulting global map is used by the navigation policy for action planning.

\subsubsection{Pose Estimator} The pose estimator is used to predict the displacement of the agent in consequence of an action. The considered atomic actions $a_t$ of the agent are: \textit{go forward 0.25m, turn left 10°, turn right 10°}. However, the noise in the actuation system and the possible physical interactions between the agent and the environment could produce unexpected outcomes causing positioning errors. The pose estimator reduces the effect of such errors by predicting the real displacement $(\Delta x_t, \Delta y_t, \Delta\theta_t)$. 
\rev{According to~\cite{ramakrishnan2020occupancy}, the input of this module consists of the RGB-D observations $(o^{rgb}_{t-1}, o^{rgb}_{t})$ and $(o^d_{t-1}, o^d_t)$ and the local maps $(l_{t-1}, l_t)$.} Each modality $i = 0, 1, 2$ is encoded singularly to obtain three different estimates of the displacement:
\begin{equation}
    g_i(e_{t-1}, e_t) = W_1 \text{max}(W_2(e_{t-1}, e_t) + b_2,0) + b_1,
    \label{eq:g_i}
\end{equation}
where $e_t \in \{o^{rgb}_t, o^d_t, l_t\}$ and $W_{1,2}$ and $b_2$ are weights matrices and bias. Eventually, the displacement estimates are aggregated with a weighted sum:
\begin{align}
    \alpha_i = \text{softmax}(\text{MLP}_i([\bar{o}^r_t, \bar{o}^d_t, \bar{l}_t])), \\
    (\Delta x_t,\Delta y_t,\Delta \theta_t) = \sum_{i=0}^{2}{\alpha_i \cdot g_i},
\label{eq:displacement}
\end{align}
where MLP is a three-layered fully-connected network, ($\bar{o}^r_t$, $\bar{o}^d_t$, $\bar{l}_t$) are the inputs encoded by a CNN, and $[\cdot, \cdot, \cdot]$ denotes tensor concatenation. The estimated pose of the agent at time $t$ is given by:
\begin{equation}
(x_t, y_t, \theta_t) = (x_{t-1}, y_{t-1}, \theta_{t-1}) + (\Delta x_t,\Delta y_t,\Delta \theta_t).
\label{eq:pose}
\end{equation}
Note that, at the beginning of each exploration episode, the agent sets its position to the center of its environment representation, \ie
\begin{equation}
(x_0, y_0, \theta_0) = (0, 0, 0).
\label{eq:initial_pose}
\end{equation}

\subsubsection{Navigation Module}
The sampling of the atomic actions of the agent relies on the hierarchical navigation policy that is composed of the following modules: the global policy, the planner, and the local policy. 

The global policy samples a point on an augmented global map of the environment, $M_t^+$, that represents the current global goal of the agent. The augmented global map $M_t^+$ is a $W\times W\times 4$ map obtained by stacking the two-channel global map $M_t$ from the Mapper with the one-hot representation of the agent position $(x_t, y_t)$ and the map of the visited positions, which collects the one-hot representations of all the positions assumed by the agent from the beginning of the exploration. 
Moreover, $M_t^+$ is in parallel cropped with respect to the position of the agent and max-pooled to a spatial dimension $H\times H$ where $H<W$. These two versions of the augmented global map are concatenated to form the $H\times H\times 8$ input of the global policy that is used to sample a goal in the global action space $H\times H$. The global policy is trained with reinforcement learning with our proposed impact-based reward $r^{global}_t$, defined below, that encourages exploration. 

The planner consists of an A* algorithm. It uses the global map to plan a path towards the global goal and samples a local goal within $1.25$m from the position of the agent. 

The local policy outputs the atomic actions
to reach the local goal and is trained to minimize the euclidean distance to the local goal, which is expressed via the following reward:
\rev{
\begin{equation}
r^{local}_t(s_t, s_{t+1}) = d(s_{t+1}) - d(s_{t}),
\label{eq:r_local}
\end{equation} 
where $d(s_t)$ is the euclidean distance to the local goal at time step $t$.}
Note that the output actions in our setup are discrete. These platform-agnostic actions can be translated into signals for specific robots actuators, as we do in this work. Alternatively, based on the high-level predicted commands, continuous actions can be predicted, \eg~in the form of linear and angular velocity commands to the robot, by using an additional, lower-level policy, as done in~\cite{irshad2021hierarchical}. The implementation of such policy is beyond the scope of our work.

Following the hierarchical structure, the global goal is reset every $\eta$ steps, and the local goal is reset if at least one of the following conditions verifies: a new global goal is sampled, the agent reaches the local goal, the local goal location is discovered to be in a occupied area.

\subsection{Impact-Driven Exploration}
The exploration ability of the agent relies on the design of an appropriate reward for the global policy. In this setting, the lack of external rewards from the environment requires the design of a dense intrinsic reward. To the best of our knowledge, our proposed method presents the first implementation of impact-driven exploration in photorealistic environments.
\rev{The key idea of this concept is encouraging the agent to perform actions that have impact on the environment and the observations retrieved from it, where the impact at time step $t$ is measured as the $l_2$-norm of the encodings of two consecutive states $\phi(s_{t})$ and $\phi(s_{t+1})$, considering the RGB observation $o_t^{rgb}$ as the state $s_{t}$. Following the formulation proposed in~\cite{raileanu2020ride}, the reward of the global policy for the proposed method is calculated as:
\begin{equation}
    r^{global}_t(s_{t}, s_{t+1}) = \frac{\left\| \phi(s_{t+1}) - \phi(s_{t}) \right\|_2}{\sqrt{N(s_{t+1})}},
\label{eq:global_reward}
\end{equation}
where $N(s_{t})$ is the visitation count of the state at time step $t$, \ie~how many times the agent has observed $s_{t}$.} The visitation count is used to drive the agent out of regions already seen in order to avoid trajectory cycles. \rev{Note that the visitation count is episodic, \ie~$N_{ep}(s_t) \equiv N(s_t)$. For simplicity, in the following we denote the episodic visitation count as $N(s_t)$.}
\subsubsection{Visitation Counts}
\rev{The concept of normalizing the reward using visitation count, as in~\cite{raileanu2020ride}, fails when the environment is continuous, since during exploration is unlikely to visit exactly the same state more than once.} In fact, even microscopic changes in terms of translation or orientation of the agent cause shifts in the values of the RGB observation, thus resulting in new states.
Therefore, using a photorealistic continuous environment nullifies the scaling property of the denominator of the global reward in Eq.~\ref{eq:global_reward} because every state $s_t$ during the exploration episode is, most of the times, only encountered for the first time. To overcome this limitation, we implement two types of pseudo-visitation counts $\hat{N}(s_t)$ to be used in place of $N(s_t)$, which extend the properties of visitation counts to continuous environments: \textit{Grid} and \textit{Density Model Estimation}.

\vspace{.1cm}

\tinytextit{Grid} With this approach, we consider a virtual discretized grid of cells with fixed size in the environment. We then assign a visitation count to each cell of the grid. 
\rev{Note that, different from approaches working on procedurally-generated environments like~\cite{raileanu2020ride}, the state space of the environment we consider is continuous also in this formulation, and depends on the pose of the agent $(x,y,\theta)$. The grid approach operates a quantization of the agent's positions, and that allows to cluster observation made from similar positions.} 
To this end, we take the global map of the environment and divide it into cells of size $G\times G$. The estimated pose of the agent, regardless of its orientation $\theta_t$, is used to select the cell that the agent occupies at time $t$. 
In the \textit{Grid} formulation, the visitation count of the selected cell is used as $N(s_t)$ in Eq.~\ref{eq:global_reward} and is formalized as:
\begin{equation}
    \hat{N}(s_t) = \hat{N}(g(x_t, y_t)),
\label{eq:pseudo_grid}
\end{equation}
where $g(\cdot)$ returns the block corresponding to the estimated position of the agent. 

\tinytextit{Density Model Estimation (DME)}
Let $\rho$ be an autoregressive density model defined over the states $s \in S$, where $S$ is the set of all possible states.
\rev{We call $\rho_n(s)$ the probability assigned by $\rho$ to the state $s$ after being trained on a sequence of states $s_1, ..., s_n$, and $\rho'_n(s)$, or recoding probability~\cite{bellemare2016unifying,ostrovski2017count}, the probability assigned by $\rho$ to $s$ after being trained on $s_1, ..., s_n, s$.}
The prediction gain $PG$ of $\rho$ describes how much the model has improved in the prediction of $s$ after being trained on $s$ itself, and is defined as
\begin{equation}
    PG_n(s) = \log \rho'_n(s) - \log \rho_n(s).
\label{eq:prediction_gain}
\end{equation}
\rev{In this work, we employ a lightweight version of Gated PixelCNN~\cite{oord2016conditional} as density model. This model is trained from scratch along with the exploration policy using the states visited during the exploration, which are fed to PixelCNN one at a time, as they are encountered.}
\rev{The weights of PixelCNN are optimized continually over all the environments. As a consequence, the knowledge of the density model is not specific for a particular environment or episode.}
To compute the input of the PixelCNN model, we transform the RGB observation $o^r_t$ to grayscale and we crop and resize it to a lower size $P\times P$. The transformed observation is quantized to $B$ bins to form the final input to the model, $s_t$. The model is trained to predict the conditional probabilities of the pixels in the transformed input image, with each pixel depending only on the previous ones following a raster scan order. The output has shape $P\times P\times B$ and each of its elements represents the probability of a pixel belonging to each of the $B$ bins. The joint distribution of the input modeled by PixelCNN is:
\begin{equation}
    p(s_t) = \prod_1^{P^2}p(\chi_i|\chi_1,...,\chi_{i-1}),
    \label{eq:pixel_cnn}
\end{equation}
where $\chi_i$ is the $i^{\textit{th}}$ pixel of the image $s_t$. $\rho$ is trained to fit $p(s_t)$ by using the negative log-likelihood loss. 

Let $\hat{n}$ be the pseudo-count total, \ie~the sum of all the visitation counts of all states during the episode. The probability and the recoding probability of $s$ can be defined as: 
\begin{align}
    \rho_n(s) &= \frac{\hat{N}_n(s)}{\hat{n}}, & \rho'_n(s) &= \frac{\hat{N}_n(s) + 1}{\hat{n} + 1}.
\label{eq:probabilities}
\end{align}
Note that, if $\rho$ is learning-positive, \ie~if $PG_n(s) > 0$ for all possible sequences $s_1,...,s_n$ and all $s \in S$, we can approximate $\hat{N}_n(s)$ as: 
\begin{equation}
    \hat{N}_n(s) = \frac{\rho_n(s)(1-\rho'_n(s))}{\rho'_n(s)-\rho_n(s)} \approx (e^{PG_n(s)} - 1)^{-1}.
\label{eq:pseudo_count}
\end{equation}

To use this approximation in Eq.~\ref{eq:global_reward}, we still need to address three problems: it does not scale with the length of the episode, the density model could be not learning-positive, and $\hat{N}_n(s)$ should be large enough to avoid the reward becoming too large regardless the goal selection. In this respect, to take into account the length of the episode, we introduce a normalizing factor $n^{-1/2}$, where $n$ is the number of steps done by the agent since the start of the episode. Moreover, to force $\rho$ to be learning-positive, we clip $PG_n(s)$ to 0 when it becomes negative.
Finally, to avoid small values at the denominator of $r^{global}_t$ (Eq.~\ref{eq:global_reward}), we introduce a lower bound of 1 to the pseudo visitation count.
The resulting definition of $\hat{N}_n(s)$ in the \textit{Density Model Estimation} formulation is:
\begin{align}
    \widetilde{PG}_n &= c \cdot n^{-1/2} \cdot (PG_n(s))_+, \\
    \hat{N}_n(s) &= \max \Big\{\Big(e^{\widetilde{PG}_n(s)}-1\Big)^{-1}, 1\Big\},
\label{eq:pseudo_count_final}
\end{align}
where $c$ is a term used to scale the prediction gain.
\rev{It is worth noting that, unlike the Grid approach that can be applied only when $s_t$ is representable as the robot location, the Density Model Estimation can be adapted to a wider range of tasks, including settings where the agent alters the environment.}

%% file: sections/04-experiments.tex
\section{Experimental Setup}
\label{sec:experiments}

\tinytit{Datasets} For comparison with state-of-the-art DRL-based methods for embodied exploration, we employ the photorealistic simulated 3D environments contained in the Gibson dataset~\cite{xia2018gibson} and the MP3D dataset~\cite{chang2017matterport3d}. Both these datasets consist of indoor environments where different exploration episodes take place. In each episode, the robot starts exploring from a different point in the environment.
Environments used during training do not appear in the validation/test split of these datasets.
Gibson contains $106$ scans of different indoor locations, for a total of around $5$M exploration episodes ($14$ locations are used in $994$ episodes for test in the so-called Gibson Val split). 
MP3D consists of $90$ scans of large indoor environments ($11$ of those are used in $495$ episodes for the validation split and $18$ in $1008$ episodes for the test split).

\tit{Evaluation Protocol}
We train our models on the Gibson train split. Then, we perform model selection basing on the results obtained on Gibson Val. We then employ the MP3D validation and test splits to benchmark the generalization abilities of the agents.
To evaluate exploration agents, we employ the following metrics. \textbf{IoU} between the reconstructed map and the ground-truth map of the environment: here we consider two different classes for every pixel in the map (free or occupied). Similarly, the map accuracy (\textbf{Acc}, expressed in $m^2$) is the portion of the map that has been correctly mapped by the agent. The area seen (\textbf{AS}, in $m^2$) is the total area of the environment observed by the agent. For both the IoU and the area seen, we also present the results relative to the two different classes: free space and occupied space respectively (\textbf{FIoU}, \textbf{OIoU}, \textbf{FAS}, \textbf{OAS}). Finally, we report the mean positioning error achieved by the agent at the end of the episode. A larger translation error (\textbf{TE}, expressed in $m$) or angular error (\textbf{AE}, in degrees) indicates that the agent struggles to keep a correct estimate of its position throughout the episode. For all the metrics, we consider episodes of length $T=500$ and $T=1000$ steps.

\rev{For our comparisons, we consider five baselines trained with different rewards. \textit{Curiosity} employs a surprisal-based intrinsic reward as defined in~\cite{pathak2017curiosity}. \textit{Coverage} and \textit{Anticipation} are trained with the corresponding coverage-based and accuracy-based rewards defined in~\cite{ramakrishnan2020occupancy}. For completeness, we include two count-based baselines, obtained using the reward defined in Eq.~\ref{eq:global_reward}, but ignoring the contribution of impact (\ie setting the numerator to a constant value of 1). These are \textit{Count (Grid)} and \textit{Count (DME)}. All the baselines share the same overall architecture and training setup of our main models.}

\input{tables/ablation}

\input{tables/tabella_brutta}

\tit{Implementation Details} The experiments are performed using the Habitat Simulator~\cite{savva2019habitat} with observations of the agent set to be $128\times 128$ RGB-D images and episode length during training set to $T=500$. Each model is trained with the training split of the Gibson dataset~\cite{xia2018gibson} with $40$ environments in parallel for $\approx5$M frames.\\
\tinytextit{Navigation Module} The reinforcement learning algorithm used to train the global and local policies is PPO~\cite{schulman2017proximal} with Adam optimizer and a learning rate of $2.5\times 10^{-4}$. The global goal is reset every $\eta=25$ time steps and the global action space hyperparameter $H$ is $240$. The local policy is updated every $\eta$ steps and the global policy is updated every $20\eta$ steps.\\
\tinytextit{Mapper and Pose Estimator} These models are trained with a learning rate of $10^{-3}$ with Adam optimizer, the local map size is set with $V=101$ while the global map size is $W=961$ for episodes in the Gibson dataset and $W=2001$ in the MP3D dataset. Both models are updated every $4\eta$ time steps, where $\eta$ is the reset interval of the global policy.\\
\tinytextit{Density Model} The model used for density estimation is a lightweight version of Gated PixelCNN~\cite{oord2016conditional} consisting of a $7\times 7$ masked convolution followed by two residual blocks with $1\times 1$ masked convolutions with $16$ output channels, a $1\times 1$ masked convolutional layer with $16$ output channels, and a final $1\times 1$ masked convolution that returns the output logits with shape $P\times P\times B$, where $B$ is the number of bins used to quantize the model input. We set $P=42$ for the resolution of the input and the output of the density model\rev{, and $c=0.1$ for the prediction gain scale factor.}

\section{Experimental Results}

\tinytit{Exploration Results}
As a first step, we perform model selection using the results on the Gibson Val split (Table~\ref{tab:ablation}). Our agents have different hyperparameters that depend on the implementation for the pseudo-counts. When our model employs grid-based pseudo-counts, it is important to determine the dimension of a single cell in this grid-based structure. In our experiments, we test the effects of using $G \times G$ squared cells, with $G \in \{ 2, 4, 5, 10 \}$. The best results are obtained with $G=5$, with small differences among the various setups.
When using pseudo-counts based on a density model, the most relevant hyperparameters depend on the particular model employed as density estimator. In our case, we need to determine the number of bins $B$ for PixelCNN, with $B\in\{64,128,256\}$. We find out that the best results are achieved with $B = 128$.
\begin{figure*}[t!]
\centering
\scriptsize
\setlength{\tabcolsep}{.3em}
\begin{tabular}{cc ccccc}
& & \textbf{Curiosity} &\textbf{Coverage} & \textbf{Anticipation} & \textbf{Impact (Grid)} & \textbf{Impact (DME)} \\
\addlinespace[0.08cm]
\rotatebox{90}{\parbox[t]{0.56in}{\hspace*{\fill}\textbf{Gibson Val}\hspace*{\fill}}} & &
\includegraphics[width=0.18\linewidth]{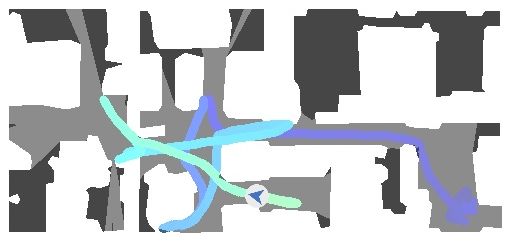} &
\includegraphics[width=0.18\linewidth]{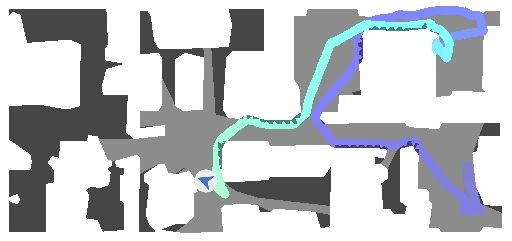} &
\includegraphics[width=0.18\linewidth]{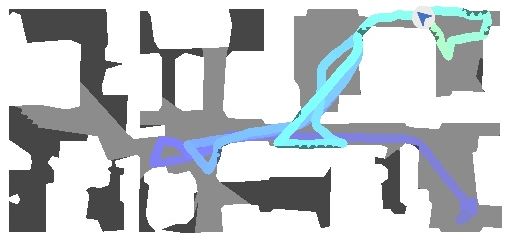} & 
\includegraphics[width=0.18\linewidth]{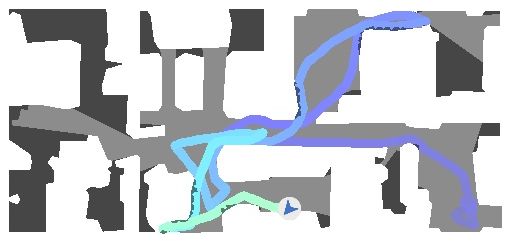} & 
\includegraphics[width=0.18\linewidth]{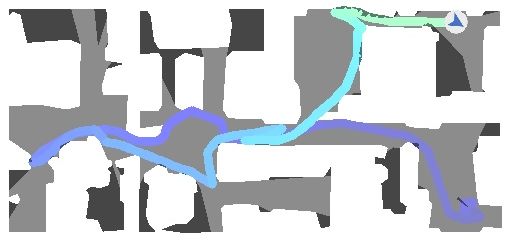} \\
\addlinespace[0.08cm]
\rotatebox{90}{\parbox[t]{1.05in}{\hspace*{\fill}\textbf{MP3D Val}\hspace*{\fill}}} & &
\includegraphics[width=0.18\linewidth]{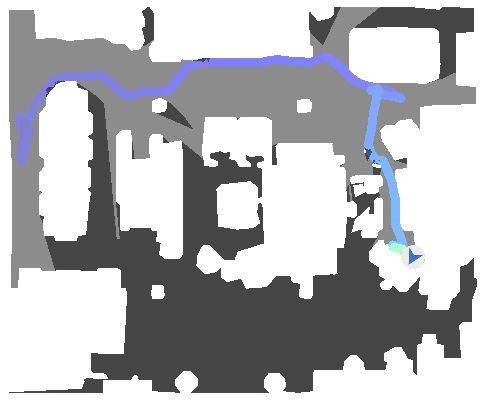} &
\includegraphics[width=0.18\linewidth]{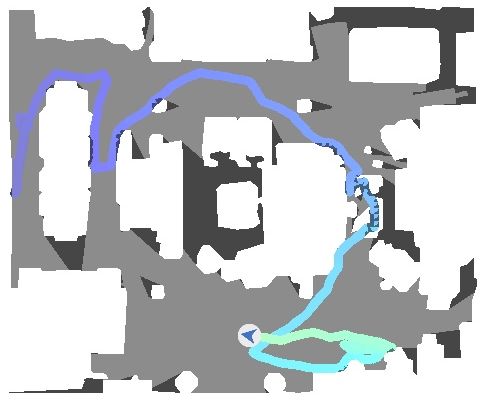} &
\includegraphics[width=0.18\linewidth]{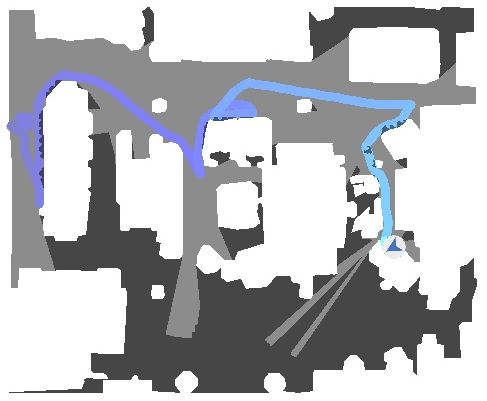} & 
\includegraphics[width=0.18\linewidth]{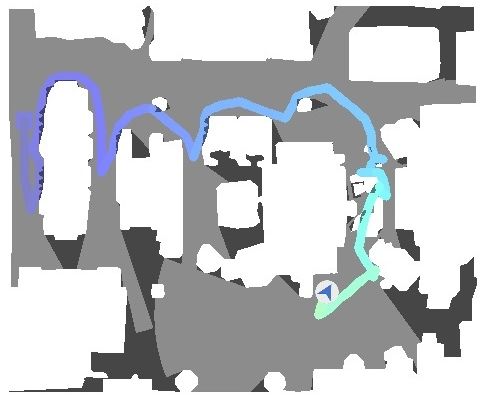} & 
\includegraphics[width=0.18\linewidth]{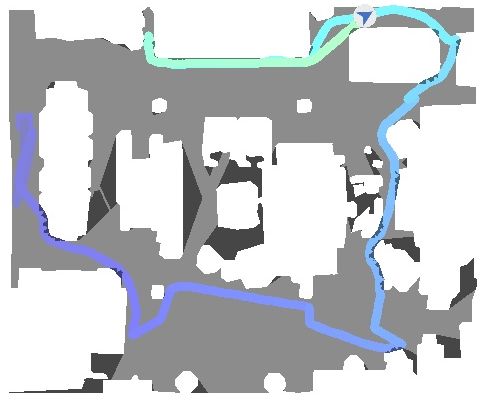} \\
\addlinespace[0.08cm]
\rotatebox{90}{\parbox[t]{0.65in}{\hspace*{\fill}\textbf{MP3D Test}\hspace*{\fill}}} & &
\includegraphics[width=0.18\linewidth]{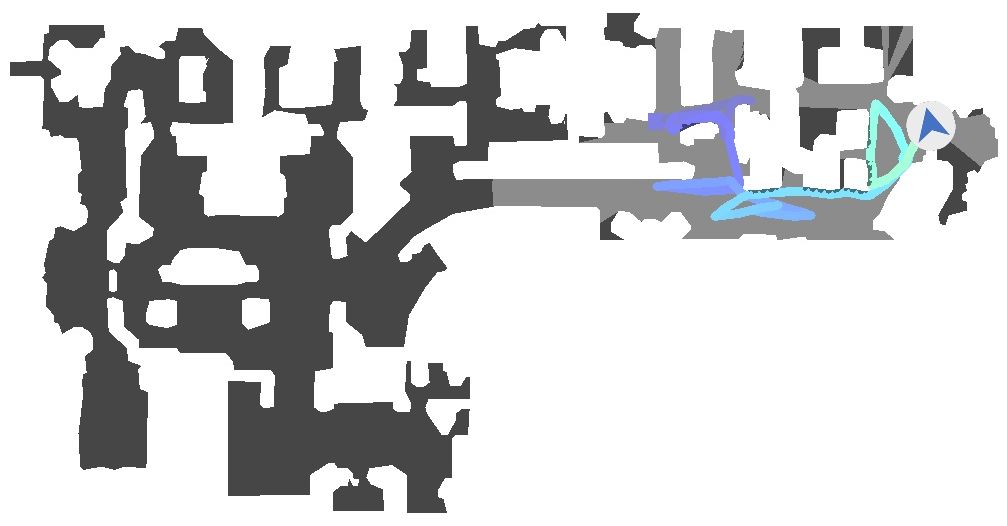} &
\includegraphics[width=0.18\linewidth]{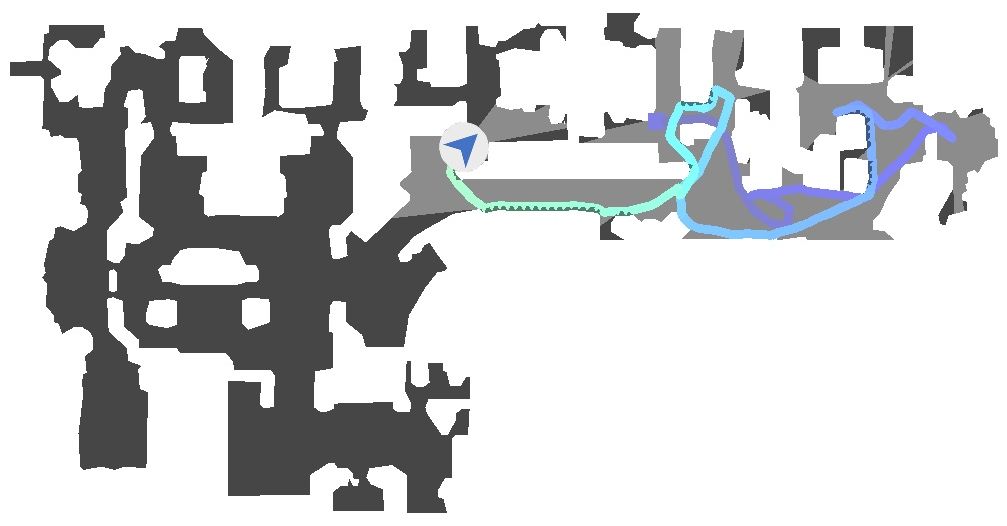} &
\includegraphics[width=0.18\linewidth]{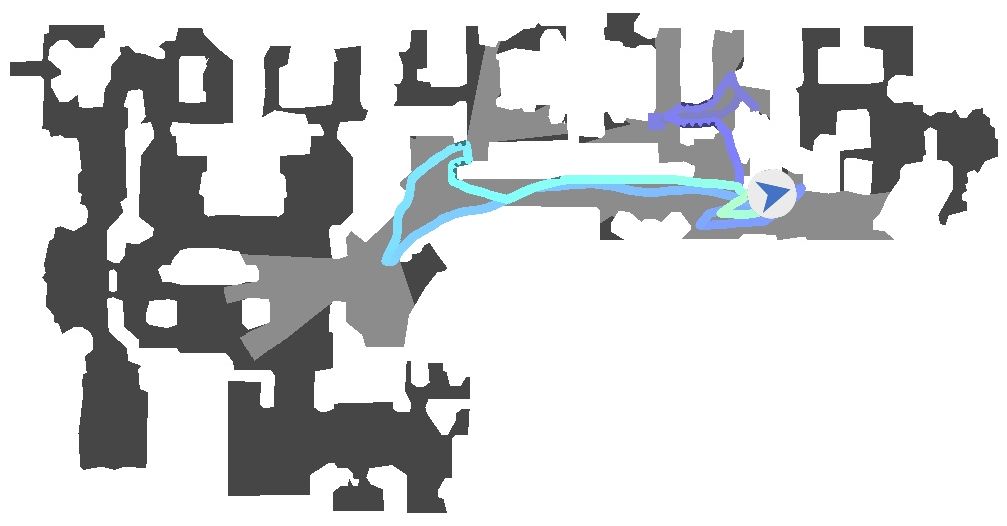} & 
\includegraphics[width=0.18\linewidth]{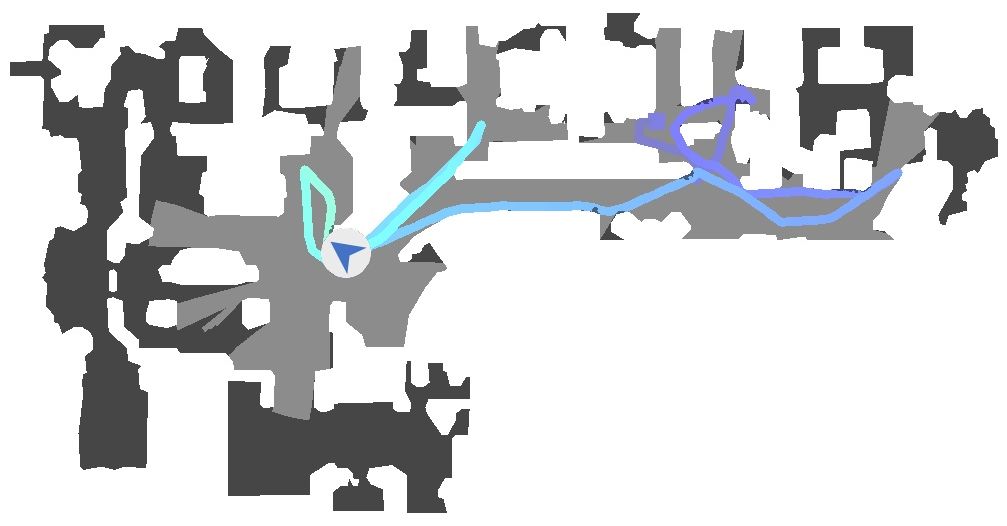} & 
\includegraphics[width=0.18\linewidth]{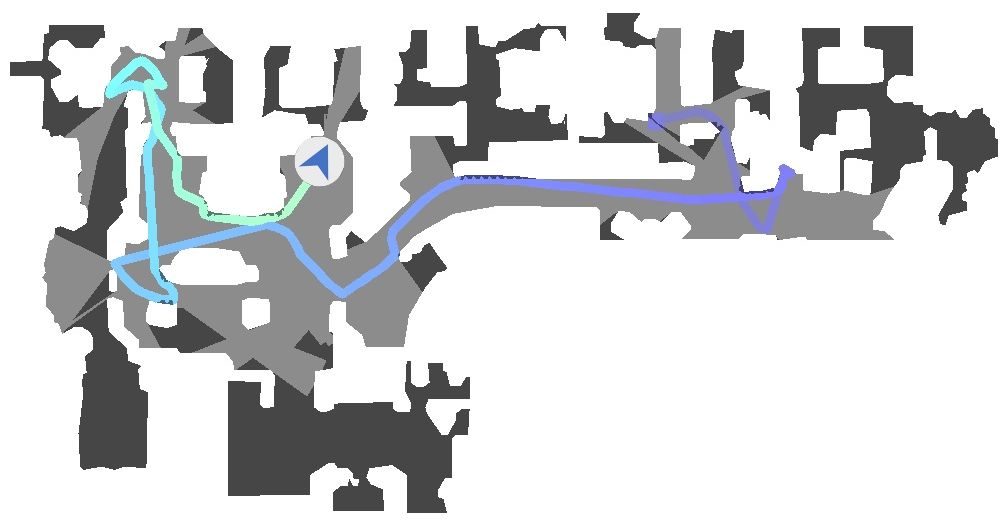} \\
\end{tabular}
\caption{Qualitative results. For each model, we report three exploration episodes on Gibson and MP3D
datasets for $T=500$.}
\label{fig:qualitatives}
\vspace{-.25cm}
\end{figure*}

\rev{In Table~\ref{tab:tabellabrutta}, we compare the Impact (Grid) and Impact (DME) agents with the baseline agents previously described on the considered datasets. For each model and each split, we test 5 different random seeds and report the mean result for each metric. For the sake of readability, we do not report the standard deviations for the different runs, which we quantify in around 1.2\% of the mean value reported.}
\rev{As it can be seen, results achieved by the two proposed impact-based agents are constantly better than those obtained by the competitors, both for $T=500$ and $T=1000$.} 
\rev{It is worth noting that our intrinsic impact-based reward outperforms strong extrinsic rewards that exploit information computed using the ground-truth layout of the environment.}
\rev{Moreover, the different implementations chosen for the pseudo-counts affect final performance, with Impact (DME) bringing the best results in terms of AS and Impact (Grid) in terms of IoU metrics.
From the results it also emerges that, although the proposed implementations for the pseudo-count in Eq.~\ref{eq:global_reward} lead to comparable results in small environments as those contained in Gibson and MP3D Val, the advantage of using DME is more evident in large, complex environments as those in MP3D Test.}

In Fig.~\ref{fig:qualitatives}, we report some qualitative results displaying the trajectories and the area seen by different agents
in the same episode. Also from a qualitative point of view, the benefit given by the proposed reward in terms of exploration trajectories and explored areas is easy to identify.

\tit{PointGoal Navigation} 
One of the main advantages of training deep modular agents for embodied exploration is that they easily adapt to perform downstream tasks, such as PointGoal navigation~\cite{savva2019habitat}. Recent literature~\cite{chaplot2019learning,ramakrishnan2020occupancy} has discovered that hierarchical agents trained for exploration are competitive with state-of-the-art architecture tailored for PointGoal navigation and trained with strong supervision for $2.5$ billion frames~\cite{wijmans2019dd}. Additionally, the training time and data required to learn the policy is much more limited (2 to 3 orders of magnitude smaller). 
\rev{In Table~\ref{tab:pointnav}, we report the results obtained using two different settings. The \textit{noise-free pose sensor} setting is the standard benchmark for PointGoal navigation in Habitat~\cite{savva2019habitat}. In the \textit{noisy pose sensor} setting, instead, the pose sensor readings are noisy, and thus the agent position must be estimated as the episode progresses.}
\rev{We consider four main metrics: the average distance to the goal achieved by the agent (\textbf{D2G}) and three success-related metrics. The success rate (\textbf{SR}) is the fraction of episodes terminated within $0.2$ meters from the goal, while the \textbf{SPL} and SoftSPL (\textbf{sSPL}) weigh the distance from the goal with the length of the path taken by the agent in order to penalize inefficient navigation.}
\rev{As it can be seen, the two proposed agents outperform the main competitors from the literature: Active Neural Slam (ANS)~\cite{chaplot2019learning} and OccAnt~\cite{ramakrishnan2020occupancy} (for which we report both the results from the paper and the official code release).}

\input{tables/pointnav_2}

\rev{When comparing with our baselines in the noise-free setting, the overall architecture design allows for high-performance results, as the reward influences map estimation only marginally. In fact, in this setting, the global policy and the pose estimation module are not used, as the global goal coincides with the episode goal coordinates, and the agent receives oracle position information. Thus, good results mainly depend on the effectiveness of the mapping module.
Instead, in the noisy setting, the effectiveness of the reward used during training influences navigation performance more significantly. In this case, better numerical results originate from a better ability to estimate the precise pose of the agent during the episode.
}
\rev{For completeness, we also compare with the results achieved by DD-PPO~\cite{wijmans2019dd}, a method trained with reinforcement learning for the PointGoal task on $2.5$ billion frames, $500$ times more than the frames used to train our agents.}

\tit{Real-world Deployment}
As agents trained in realistic indoor environments using the Habitat simulator are adaptable to real-world deployment~\cite{kadian2020sim2real,bigazzi2021out}, we also deploy the proposed approach on a LoCoBot robot\footnote{\url{https://locobot-website.netlify.com}}. We employ the PyRobot interface~\cite{murali2019pyrobot} to deploy code and trained models on the robot.
To enable the adaptation to the real-world environment, there are some aspects that must be taken into account during training. As a first step, we adjust the simulation in order to reproduce realistic actuation and sensor noise. To that end, we adopt the noise model proposed in~\cite{chaplot2019learning} based on Gaussian Mixture Models fitting real-world noise data acquired from a LoCoBot. Additionally, we modify the parameters of the RGB-D sensor used in simulation to match those of the RealSense camera mounted on the robot. Specifically, we change the camera resolution and field of view, the range of depth information, and the camera height. Finally, it is imperative to prevent the agent from learning simulation-specific shortcuts and tricks. For instance, the agent may learn to slide along the walls due to imperfect dynamics in simulation~\cite{kadian2020sim2real}. To prevent the learning of such dynamics, we employ the \textit{bump} sensor provided by Habitat and block the agent whenever it is in contact with an obstacle.
When deployed in the real world, our agent is able to explore the environment without getting stuck or bumping into obstacles.
In the video accompanying the submission, we report exploration samples taken from our real-world deployment.

%% file: tables/ablation.tex
\begin{table}[t]
\footnotesize
\centering
\caption{Results for our model selection on Gibson Val for $T=500$.}
\label{tab:ablation}
\setlength{\tabcolsep}{.35em}
\resizebox{\linewidth}{!}{
\begin{tabular}{l c ccccccccc}
\toprule
% & & \multicolumn{9}{c}{\textbf{Gibson Val (T = 500)}} \\
% \cmidrule{3-11}
\textbf{Model} & & \textbf{IoU} $\uparrow$ & \textbf{FIoU} $\uparrow$ & \textbf{OIoU} $\uparrow$ & \textbf{Acc} $\uparrow$ & \textbf{AS} $\uparrow$ & \textbf{FAS} $\uparrow$ & \textbf{OAS} $\uparrow$ & \textbf{TE} $\downarrow$ & \textbf{AE} $\downarrow$ \\
%  & & \multicolumn{3}{c}{\textbf{IoU}} & & \textbf{MapAcc} & & \multicolumn{3}{c}{\textbf{AreaSeen}} & & \multicolumn{2}{c}{\textbf{PosError}} \\
% \cmidrule{3-5} \cmidrule{7-7} \cmidrule{9-11} \cmidrule{13-14}
% \textbf{Model} & & \textbf{Tot} & \textbf{F} & \textbf{O} & & \textbf{Tot} & & \textbf{Tot} & \textbf{F} & \textbf{O} & & \textbf{Trans} & \textbf{Ang} \\
\midrule
% \midrule
% \textbf{\textit{Coverage}} & & 0.717 & 0.711 & 0.723 & 50.97 & 60.50 & 33.77 & 26.73 & 0.270 & 5.499 \\
% \textbf{\textit{Anticipation}} & & 0.782 & 0.777 & 0.789 & 54.44 & 60.70 & 33.90 & 26.80 & 0.101 & 1.097 \\
% \textbf{\textit{Curiosity}} & & 0.678 & 0.668 & 0.687 & 49.24 & 61.45 & 33.99 & 27.46 & 0.328 & 7,411 \\
% \midrule
\textbf{Grid} \\
\textit{  G = 2} & & 0.726 & 0.721 & 0.730 & 51.41 & 61.88 & 34.17 & 27.71 & 0.240 & 4.450 \\
\textit{  G = 4} & & 0.796 & 0.792 & 0.801 & 54.34 & 61.17 & 33.74 & 27.42 & 0.079 & 1.055 \\
\textit{  G = 5} & & \textbf{0.806} & \textbf{0.801} & \textbf{0.813} & \textbf{55.21} & \textbf{62.17} & \textbf{34.31} & \textbf{27.87} & \textbf{0.077} & \textbf{0.881} \\
\textit{  G = 10} & & 0.789 & 0.784 & 0.794 & 54.26 & 61.67 & 34.06 & 27.61 & 0.111 & 1.434 \\
\midrule
\textbf{DME} \\
\textit{  B = 64} & & 0.773 & 0.768 & 0.778 & 53.58 & 61.00 & 33.79 & 27.21 & 0.131 & 2.501 \\
\textit{  B = 128} & & \textbf{0.796} & \textbf{0.794} & \textbf{0.799} & \textbf{54.73} & \textbf{62.07} & \textbf{34.27} & \textbf{27.79} & \textbf{0.095} & \textbf{1.184} \\
\textit{  B = 256} & & 0.685 & 0.676 & 0.695 & 49.27 & 61.40 & 33.95 & 27.45 & 0.311 & 6.817 \\
\bottomrule
\end{tabular}
}
\vspace{-0.3cm}
\end{table}

%% file: tables/tabella_brutta.tex
\begin{table*}[t]
\footnotesize
\centering
\caption{Exploration results on Gibson Val, MP3D Val, and MP3D Test, at different episode maximum length.}
\label{tab:tabellabrutta}
\setlength{\tabcolsep}{.32em}
\resizebox{\linewidth}{!}{
%\begin{tabular}{lc ccccccccc c ccccccccc}
\begin{tabular}{>{\color{black}}l>{\color{black}}c >{\color{black}}c>{\color{black}}c>{\color{black}}c>{\color{black}}c>{\color{black}}c>{\color{black}}c>{\color{black}}c>{\color{black}}c>{\color{black}}c>{\color{black}}c >{\color{black}}c>{\color{black}}c>{\color{black}}c>{\color{black}}c>{\color{black}}c>{\color{black}}c>{\color{black}}c>{\color{black}}c>{\color{black}}c}
\toprule
& & \multicolumn{9}{>{\color{black}}c}{\textbf{Gibson Val (T = 500)}}
& & \multicolumn{9}{>{\color{black}}c}{\textbf{Gibson Val (T = 1000)}} \\
\cmidrule{3-11} \cmidrule{13-21}
\textbf{Model} & & \textbf{IoU} $\uparrow$ & \textbf{FIoU} $\uparrow$ & \textbf{OIoU} $\uparrow$ & \textbf{Acc} $\uparrow$ & \textbf{AS} $\uparrow$ & \textbf{FAS} $\uparrow$ & \textbf{OAS} $\uparrow$ & \textbf{TE} $\downarrow$ & \textbf{AE} $\downarrow$
& & \textbf{IoU} $\uparrow$ & \textbf{FIoU} $\uparrow$ & \textbf{OIoU} $\uparrow$ & \textbf{Acc} $\uparrow$ & \textbf{AS} $\uparrow$ & \textbf{FAS} $\uparrow$ & \textbf{OAS} $\uparrow$ & \textbf{TE} $\downarrow$ & \textbf{AE} $\downarrow$ \\
\midrule
Curiosity
& & 0.678 & 0.669 & 0.688 & 49.35 & 61.67 & 34.16 & 27.51 & 0.330 & 7.430
& & 0.560 & 0.539 & 0.581 & 45.71 & 67.64 & 37.19 & 30.45 & 0.682 & 14.862 \\
Coverage 
& & 0.721 & 0.715 & 0.726 & 51.47 & 61.13 & 34.07 & 27.06 & 0.272 & 5.508
& & 0.653 & 0.641 & 0.664 & 50.10 & 66.15 & 36.77 & 29.38 & 0.492 & 10.796 \\
Anticipation 
& & 0.783 & 0.778 & 0.789 & 54.68 & 60.96 & 34.15 & 26.81 & 0.100 & 1.112
& & 0.773 & 0.763 & 0.782 & 56.37 & 66.61 & 37.17 & 29.44 & 0.155 & 1.876 \\
\midrule
\rev{Count (Grid)}
& & \rev{0.714} & \rev{0.706} & \rev{0.721} & \rev{50.85} & \rev{61.61} & \rev{34.17} & \rev{27.44} & \rev{0.258} & \rev{5.476}
& & \rev{0.608} & \rev{0.592} & \rev{0.624} & \rev{48.22} & \rev{67.80} & \rev{37.31} & \rev{30.50} & \rev{0.520} & \rev{10.996} \\
\rev{Count (DME)}
& & \rev{0.764} & \rev{0.757} & \rev{0.772} & \rev{52.81} & \rev{60.69} & \rev{33.68} & \rev{27.01} & \rev{0.148} & \rev{2.888}
& & \rev{0.708} & \rev{0.694} & \rev{0.722} & \rev{52.67} & \rev{66.91} & \rev{36.81} & \rev{30.12} & \rev{0.282} & \rev{5.802} \\
\midrule
\textbf{Impact (Grid)}
& & \textbf{0.803} & \textbf{0.797} & \textbf{0.809} &          54.94 & 61.90 & 34.07 & 27.83 & \textbf{0.079} & \textbf{0.878}
& & \textbf{0.802} & \textbf{0.793} & \textbf{0.811} & \textbf{57.21} & 67.74 & 37.04 & 30.69 & \textbf{0.119} & \textbf{1.358} \\
% \midrule
\textbf{Impact (DME)}
& & 0.800 & 0.796 & 0.803 & \textbf{55.10} & \textbf{62.59} & \textbf{34.45} & \textbf{28.14} & 0.095 & 1.166
& & 0.789 & 0.783 & 0.796 &          56.77 & \textbf{68.34} & \textbf{37.42} & \textbf{30.92} & 0.154 & 1.958 \\
\midrule
\addlinespace[0.1cm]
\midrule
& & \multicolumn{9}{>{\color{black}}c}{\textbf{MP3D Val (T = 500)}} & & \multicolumn{9}{>{\color{black}}c}{\textbf{MP3D Val (T = 1000)}} \\
\cmidrule{3-11} \cmidrule{13-21}
\textbf{Model} & & \textbf{IoU} $\uparrow$ & \textbf{FIoU} $\uparrow$ & \textbf{OIoU} $\uparrow$ & \textbf{Acc} $\uparrow$ & \textbf{AS} $\uparrow$ & \textbf{FAS} $\uparrow$ & \textbf{OAS} $\uparrow$ & \textbf{TE} $\downarrow$ & \textbf{AE} $\downarrow$
& & \textbf{IoU} $\uparrow$ & \textbf{FIoU} $\uparrow$ & \textbf{OIoU} $\uparrow$ & \textbf{Acc} $\uparrow$ & \textbf{AS} $\uparrow$ & \textbf{FAS} $\uparrow$ & \textbf{OAS} $\uparrow$ & \textbf{TE} $\downarrow$ & \textbf{AE} $\downarrow$ \\
\midrule
Curiosity
& & 0.339 & 0.473 & 0.205 & 97.82 & 118.13 & 75.73 & 42.40 & 0.566 & 7.290
& & 0.336 & 0.449 & 0.223 & 109.79 & 157.27 & 100.07 & 57.20 & 1.322 & 14.540 \\
Coverage 
& & 0.352 & 0.494 & 0.210 & 102.05 & 120.00 & 76.78 & 43.21 & 0.504 & 5.822
& & 0.362 & 0.492 & 0.232 & 116.58 & 158.83 & 100.76 & 58.07 & 1.072 & 11.624 \\
Anticipation 
& & 0.381 & 0.530 & 0.231 & 106.02 & 114.06 & 72.94 & 41.13 & 0.151 & 1.280
& & 0.420 & 0.568 & 0.272 & 126.86 & 147.33 & 93.56 & 53.78 & 0.267 & 2.436 \\
\midrule
\rev{Count (Grid)}
& & \rev{0.347} & \rev{0.488} & \rev{0.206} & \rev{99.00} & \rev{116.77} & \rev{75.00} & \rev{41.76} & \rev{0.466} & \rev{5.828} 
& & \rev{0.350} & \rev{0.474} & \rev{0.226} & \rev{112.75} & \rev{157.13} & \rev{100.03} & \rev{57.10} & \rev{1.074} & \rev{11.686} \\
\rev{Count (DME)}
& & \rev{0.359} & \rev{0.493} & \rev{0.225} & \rev{101.73} & \rev{112.65} & \rev{72.22} & \rev{40.43} & \rev{0.268} & \rev{3.318} 
& & \rev{0.379} & \rev{0.505} & \rev{0.254} & \rev{119.07} & \rev{149.62} & \rev{95.16} & \rev{54.46} & \rev{0.590} & \rev{6.544} \\
\midrule
\textbf{Impact (Grid)} 
& &          0.383 &          0.531 & \textbf{0.234} &          107.41 & 116.60 & 74.44 & 42.17 & \textbf{0.120} & \textbf{0.860}
& & \textbf{0.440} & \textbf{0.595} & \textbf{0.285} & \textbf{133.97} & 157.19 & 99.61 & 57.58 & \textbf{0.202} & \textbf{1.294} \\
\textbf{Impact (DME)}
& & \textbf{0.396} & \textbf{0.560} & 0.233 & \textbf{111.61} & \textbf{124.06} &  \textbf{79.47} & \textbf{44.59} & 0.232 & 1.988
& &          0.427 &          0.587 & 0.268 &          133.27 & \textbf{166.20} & \textbf{105.69} & \textbf{60.50} & 0.461 & 3.654 \\
\midrule
\addlinespace[0.1cm]
\midrule
& & \multicolumn{9}{>{\color{black}}c}{\textbf{MP3D Test (T = 500)}}
& & \multicolumn{9}{>{\color{black}}c}{\textbf{MP3D Test (T = 1000)}} \\
\cmidrule{3-11} \cmidrule{13-21}
\textbf{Model} & & \textbf{IoU} $\uparrow$ & \textbf{FIoU} $\uparrow$ & \textbf{OIoU} $\uparrow$ & \textbf{Acc} $\uparrow$ & \textbf{AS} $\uparrow$ & \textbf{FAS} $\uparrow$ & \textbf{OAS} $\uparrow$ & \textbf{TE} $\downarrow$ & \textbf{AE} $\downarrow$
& & \textbf{IoU} $\uparrow$ & \textbf{FIoU} $\uparrow$ & \textbf{OIoU} $\uparrow$ & \textbf{Acc} $\uparrow$ & \textbf{AS} $\uparrow$ & \textbf{FAS} $\uparrow$ & \textbf{OAS} $\uparrow$ & \textbf{TE} $\downarrow$ & \textbf{AE} $\downarrow$ \\
\midrule
Curiosity
& & 0.362 & 0.372 & 0.352 & 109.66 & 130.48 & 85.98 & 44.50 & 0.620 & 7.482
& & 0.361 & 0.365 & 0.357 & 130.10 & 185.36 & 121.65 & 63.71 & 1.520 & 14.992 \\
Coverage 
& & 0.390 & 0.401 & 0.379 & 116.71 & 134.89 & 88.15 & 46.75 & 0.564 & 5.938
& & 0.409 & 0.418 & 0.399 & 142.86 & 193.20 & 126.21 & 66.99 & 1.240 & 11.814 \\
Anticipation 
& & 0.424 & 0.433 & \textbf{0.415} & 117.87 & 124.24 & 81.31 & 42.93 & 0.151 & 1.306 
& & 0.484 & 0.491 & 0.478 & 153.83 & 174.76 & 114.29 & 60.47 & 0.289 & 2.356 \\
\midrule
Count (Grid)
& & 0.364 & 0.381 & 0.348 & 117.50 & 134.85 & 89.81 & 45.05 & 0.525 & 5.790 
& & 0.377 & 0.391 & 0.363 & 144.26 & 194.76 & 129.22 & 65.53 & 1.246 & 11.608 \\
Count (DME)
& & 0.391 & 0.397 & 0.385 & 114.02 & 123.86 & 81.86 & 42.00 & 0.287 & 3.322 
& & 0.418 & 0.419 & 0.418 & 140.21 & 172.44 & 113.25 & 59.19 & 0.657 & 6.572 \\
\midrule
\textbf{Impact (Grid)}
& & 0.420 & 0.430 & 0.409 & 124.44 & 130.98 & 86.08 & 44.90 & \textbf{0.124} & \textbf{0.834} 
& & \textbf{0.502} & \textbf{0.510} & \textbf{0.494} & 168.55 & 190.03 & 124.44 & 65.60 & \textbf{0.218} & \textbf{1.270} \\
\textbf{Impact (DME)}
& & \textbf{0.426} & \textbf{0.444} & 0.409 & \textbf{133.51} & \textbf{144.64} & \textbf{95.70}  & \textbf{48.94} & 0.288 & 2.312
& &          0.481 &          0.498 &          0.464 & \textbf{174.18} & \textbf{212.00} & \textbf{140.10} & \textbf{71.90} & 0.637 & 4.390 \\
\bottomrule
\end{tabular}
}
\vspace{-0.2cm}
\end{table*}

%% file: tables/pointnav_2.tex
\begin{table}[t]
\footnotesize
\centering
\caption{PointGoal Navigation results on the Validation subset of the Gibson dataset. Underlined denotes second best. % DD-PPO~\cite{wijmans2019dd} is trained specifically for PointGoal Navigation.
} 
\label{tab:pointnav}
\setlength{\tabcolsep}{.3em}
\resizebox{\linewidth}{!}{
% \begin{tabular}{lc cccc c cccc}
\begin{tabular}{lc cccc c >{\color{black}}c >{\color{black}}c >{\color{black}}c >{\color{black}}c}
\toprule
& & \multicolumn{4}{c}{\textbf{Noise-free Pose Sensor}} & & \multicolumn{4}{c}{\rev{\textbf{Noisy Pose Sensor}}} \\
\cmidrule{3-6} \cmidrule{8-11} 
\textbf{Model} & & \textbf{D2G} $\downarrow$ & \textbf{SR} $\uparrow$ & \textbf{SPL} $\uparrow$ & \textbf{sSPL} $\uparrow$ & & \textbf{D2G} $\downarrow$ & \textbf{SR} $\uparrow$ & \textbf{SPL} $\uparrow$ & \textbf{sSPL} $\uparrow$ \\ 
\midrule
ANS~\cite{chaplot2019learning} & & - & 0.950 & 0.846 & - & & - & - & - & - \\
OccAnt~\cite{ramakrishnan2020occupancy} & & - & 0.930 & 0.800 & - & & - & - & - & - \\
OccAnt~\cite{ramakrishnan2020occupancy}\tablefootnote{\url{https://github.com/facebookresearch/OccupancyAnticipation}} & & - & - & \underline{0.911} & - & & - & - & - & - \\
\midrule
\rev{Curiosity} & & \rev{\textbf{0.238}} & \rev{\textbf{0.970}} & \rev{\textbf{0.914}} & \rev{\textbf{0.899}} & & 0.302 & 0.861 & 0.822 & \underline{0.890} \\
\rev{Coverage} & & \rev{\underline{0.240}} & \rev{\textbf{0.970}} & \rev{0.909} & \rev{\underline{0.895}} & & 0.288 & 0.827 & 0.788 & 0.886 \\
\rev{Anticipation} & & \rev{0.285} & \rev{0.965} & \rev{0.906} & \rev{0.892} & & 0.309 & 0.885 & 0.835 & 0.884 \\
\midrule
\textbf{Impact (Grid)} & & 0.252 & \underline{0.969} & 0.908 & 0.894 & & \textbf{0.226} & \textbf{0.923} & \textbf{0.867} & \textbf{0.893} \\ 
\textbf{Impact (DME)}  & & 0.264 & 0.967 & 0.907 & \underline{0.895} & & \underline{0.276} & \underline{0.913} & \underline{0.859} & \textbf{0.893} \\ 
\midrule
\textit{DD-PPO}~\cite{wijmans2019dd} & & - & \textit{0.967} & \textit{0.922} & - & & - & - & - & - \\
\bottomrule
\end{tabular}
}
\vspace{-0.25cm}
\end{table}

%% file: sections/05-conclusion.tex
\section{Conclusion}
\label{sec:conclusion}
In this work, we presented an impact-driven approach for robotic exploration in indoor environments. Different from previous research that considered a setting with procedurally-generated environments with a finite number of possible states, we tackle a problem where the number of possible states is non-numerable. To deal with this scenario, we exploit a deep neural density model to compute a running pseudo-count of past states and use it to regularize the impact-based reward signal. \rev{The resulting intrinsic reward allows to efficiently train an agent for exploration even in absence of an extrinsic reward. Furthermore, extrinsic rewards and our proposed reward can be jointly used to improve training efficiency in reinforcement learning. The proposed agent stands out from the recent literature on embodied exploration in photorealistic environments.
Additionally,
we showed that the trained models can be deployed in the real world.}